\documentclass[10pt,twocolumn,letterpaper]{article}

\usepackage{cvpr}              

\usepackage{graphicx}
\usepackage{multirow}
\usepackage{booktabs}
\usepackage{caption}
\usepackage{colortbl}
\usepackage{xcolor}
\usepackage{bbding}
\usepackage{ulem}
\usepackage{tikz}     
\usepackage{algorithm}
\usepackage{algpseudocode}
\usepackage{amsmath}  
\usepackage[accsupp]{axessibility}
\usepackage[dvipsnames]{xcolor}

\definecolor{cvprblue}{rgb}{0.21,0.49,0.74}
\usepackage[pagebackref,breaklinks,colorlinks,allcolors=cvprblue]{hyperref}
\usepackage{float}

\def\paperID{2664} 
\def\confName{CVPR}
\def\confYear{2025}

\title{VEU-Bench: Towards Comprehensive Understanding of Video Editing}

\author{
Bozheng Li$^{1,2}$
\quad Yongliang Wu$^{1,3}$
\quad Yi Lu$^{1,4}$
\quad Jiashuo Yu$^{5}$ 
\quad Licheng Tang$^{1}$
\quad Jiawang Cao$^{1}$ \\
\quad Wenqing Zhu$^{1}$
\quad Yuyang Sun$^{1}$ 
\quad Jay Wu$^{1}$
\quad Wenbo Zhu$^{1}$ \\
$^{1}$Opus AI Research \quad\\ $^{2}$Brown University \quad $^{3}$Southeast University \quad $^{4}$University of Toronto \quad $^{5}$Fudan University \quad\\
\texttt{bozheng\_li@brown.edu} \quad \texttt{wenbo.zhu@berkeley.edu} \\
}

\begin{document}
\twocolumn[{%
    \maketitle
    \begin{figure}[H]
        \hsize=\textwidth 
        \centering
        \includegraphics[width=2\linewidth]{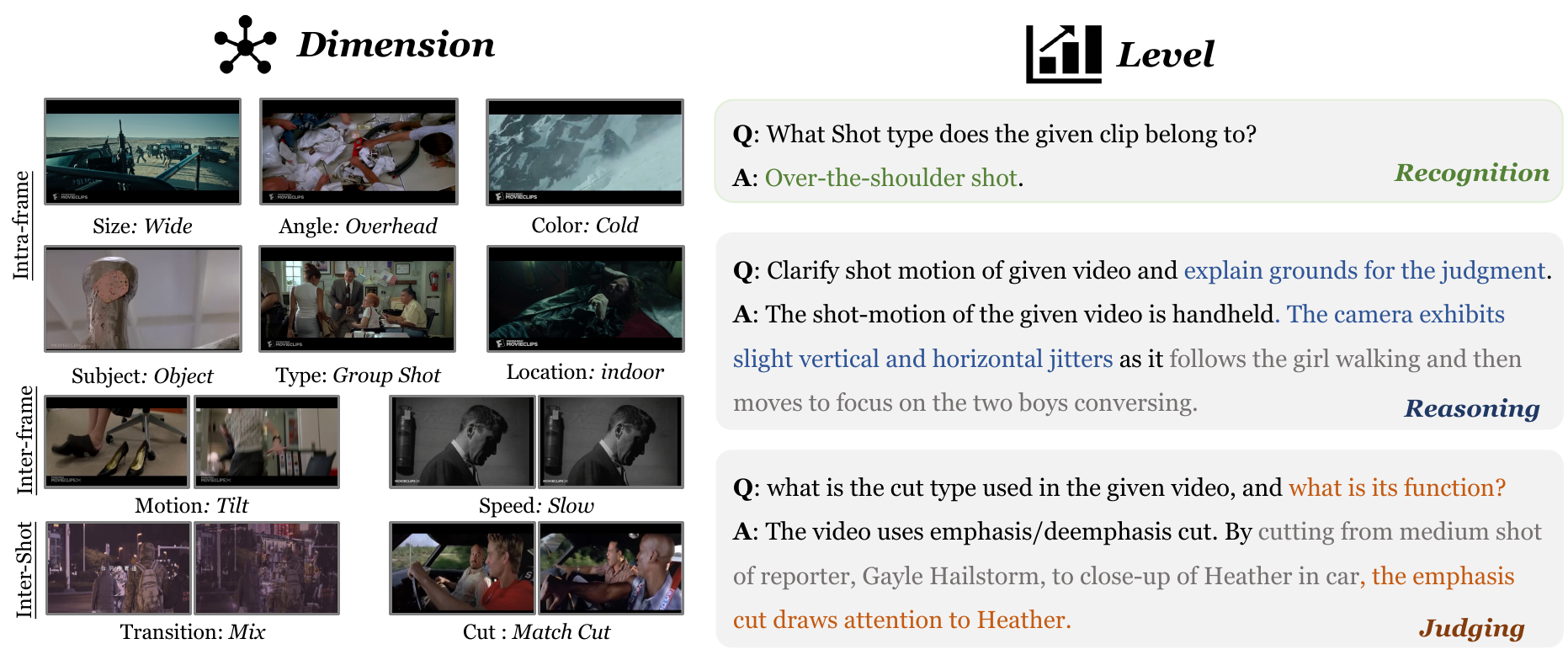}
        \caption{\textbf{The overview of our proposed VEU-Bench}. VEU-Bench covers 10 editing dimensions, evaluating models on tasks ranging from recognition to reasoning and judging, providing a robust evaluation of video editing understanding across various aspects and levels of difficulty.}
        \label{fig:teaser}
    \end{figure}
}]

\begin{abstract}
Widely shared videos on the internet are often edited. Recently, although Video Large Language Models (Vid-LLMs) have made great progress in general video understanding tasks, their capabilities in video editing understanding (VEU) tasks remain unexplored. To address this gap, in this paper, we introduce VEU-Bench (\textbf{V}ideo \textbf{E}diting \textbf{U}nderstanding \textbf{Bench}mark), a comprehensive benchmark that categorizes video editing components across various dimensions, from intra-frame features like shot size to inter-shot attributes such as cut types and transitions. Unlike previous video editing understanding benchmarks that focus mainly on editing element classification, VEU-Bench encompasses 19 fine-grained tasks across three stages: recognition, reasoning, and judging. To enhance the annotation of VEU automatically, we built an annotation pipeline integrated with an ontology-based knowledge base. Through extensive experiments with 11 state-of-the-art Vid-LLMs, our findings reveal that current Vid-LLMs face significant challenges in VEU tasks, with some performing worse than random choice. To alleviate this issue, we develop Oscars\footnote{Named after the Academy Awards.}, a VEU expert model fine-tuned on the curated VEU-Bench dataset. It outperforms existing open-source Vid-LLMs on VEU-Bench by over 28.3\% in accuracy and achieves performance comparable to commercial models like GPT-4o. We also demonstrate that incorporating VEU data significantly enhances the performance of Vid-LLMs on general video understanding benchmarks, with an average improvement of 8.3\% across nine reasoning tasks. The code and data are available at \href{https://labazh.github.io/VEU-Bench.github.io/}{project page}
\end{abstract}

\section{Introduction}
\label{sec:intro}
Nowadays large volumes of videos circulating on the internet are edited videos. Video editing involves processing and combining raw footage~\cite{metz1991film}. Video Editing Understanding (VEU) focuses on identifying and interpreting editing elements in videos, including ‘nouns’ like shot attributes~\cite{argaw2022anatomy} and ‘verbs’ such as cuts~\cite{chen2023match, pardo2022moviecuts} and transitions~\cite{shen2022autotransition, gu2024edit3k}.

The value of VEU tasks lies in two main areas. First, understanding editing elements enables assisted automatic editing~\cite{frey2021automatic, smith2017harnessing}, improving efficiency and allowing quality assessment of edited videos~\cite{xu2024beyond}. VEU also lowers the barrier for beginners learning video editing. Second, VEU data strengthens the ability to reason abstractly of video models. Editing elements are abstract concepts~\cite{metz1991film, dancyger2018technique} derived from specialized techniques, not directly present in the real world. For example, understanding a \textit{“Match Cut”} requires recognizing visual patterns, like shape or movement alignment, across scenes. VEU demands knowledge of editing patterns, as such abstractions aren’t readily observable in the real world, making VEU highly suitable as abstract reasoning data for video models.

Since VEU benchmarks~\cite{argaw2022anatomy, shen2022autotransition, pardo2022moviecuts} have primarily focused on classification, the reasoning and interpretation of video editing components remain underexplored. With the recent advancements in Video Large Language Models(Vid-LLMs)\cite{cheng2024videollama, fei2024video} capable of performing high-level causal reasoning on video inputs, there is strong potential for utilization of Vid-LLMs in VEU tasks. This makes a more comprehensive and challenging VEU benchmark increasingly indispensable for thoroughly evaluating the VEU ability of Vid-LLMs.

To fill this gap, we propose VEU-Bench (\textbf{V}ideo \textbf{E}diting \textbf{U}nderstanding \textbf{Bench}mark), a comprehensive benchmark that categorizes video editing components across various dimensions, from intra-frame features like shot size~\cite{argaw2022anatomy} to inter-shot attributes like cut types~\cite{pardo2022moviecuts} and transitions~\cite{shen2022autotransition}. Unlike previous benchmarks~\cite{argaw2022anatomy, shen2022autotransition, pardo2022moviecuts, chen2023match, gu2024edit3k} which focus mainly on editing element classification, VEU-Bench covers 19 fine-grained tasks across three stages: recognition, reasoning, and judging, as illustrated in Figure~\ref{fig:teaser}.

For benchmark construction, we curate video editing data from previous studies~\cite{argaw2022anatomy, shen2022autotransition, pardo2022moviecuts} and, guided by professional editing tutorials~\cite{metz1991film, wohl2002editing}, organize features and functions of editing elements into a detailed knowledge base. Using this knowledge base, we develop an ontology-based annotation pipeline~\cite{khurana2013study} to rewrite abstract editing features, such as \textit{“Vertical camera movement without changing base position”}, into video-specific descriptions like \textit{“The camera tilts vertically upward from the woman’s mouth to her eyes, with no lateral movement”}. Similarly, for judging tasks, functions are rewritten in a video-specific context. Through this pipeline, we extend VEU tasks from classification to reasoning and judging, introducing VEU-50K, a well-crafted dataset with around 50k VEU samples. Additionally, the proposed knowledge base provides domain-specific knowledge to aid LLMs in evaluating responses to open-ended questions.

\begin{figure}[t]
    \centering
    \includegraphics[width=\linewidth]{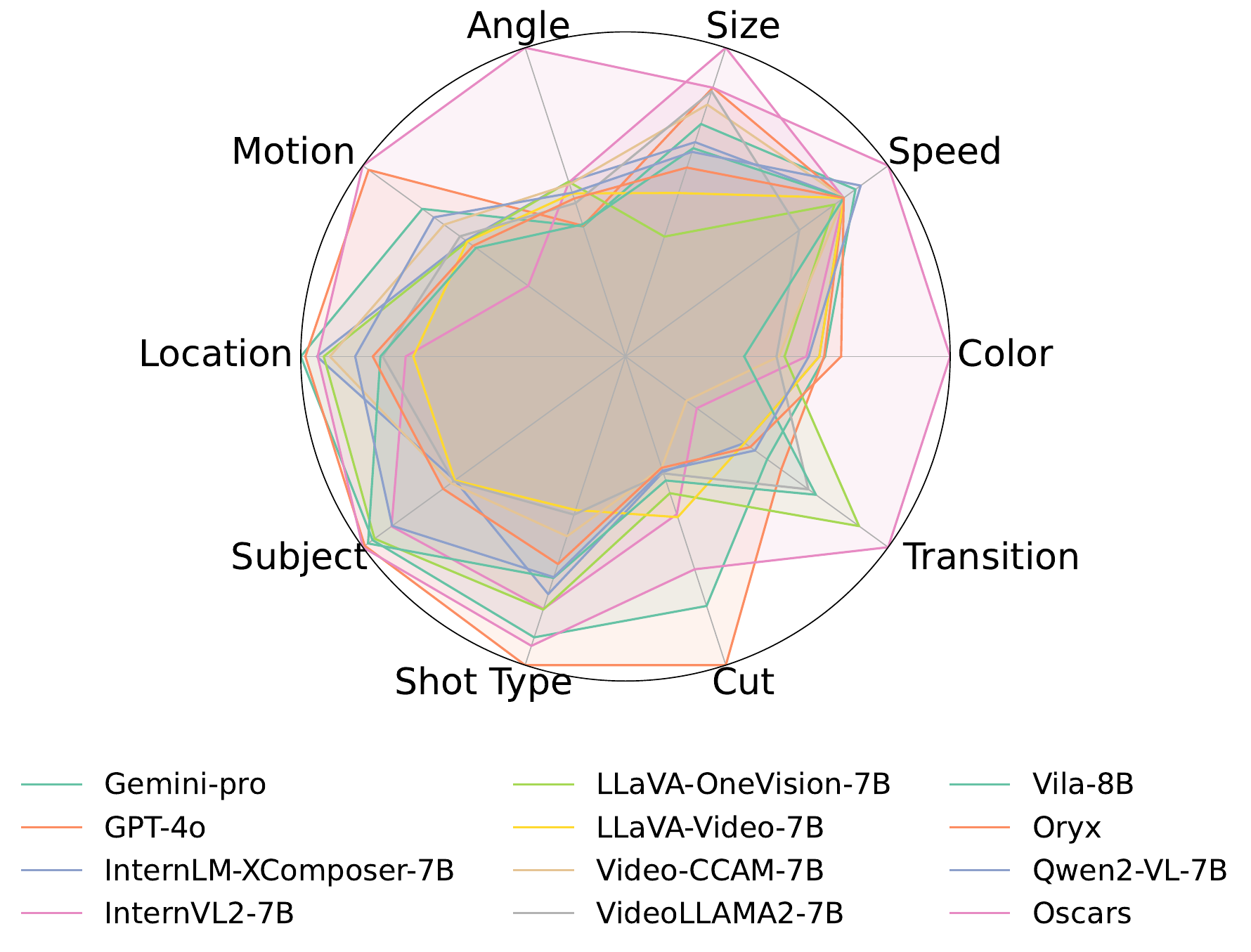}
    \caption{\textbf{The performance of 11 Vid-LLMs and the proposed expert model, Oscars, on VEU-Bench.} We normalize the results per dimension for clearer comparisons.}
    \label{fig:rader_map}
\end{figure}

We conduct extensive evaluations of 11 Vid-LLMs on VEU-Bench and found that current state-of-the-art models struggle to understand editing components and patterns, as shown in Figure~\ref{fig:rader_map}. Therefore, we developed Oscars, an expert Vid-LLM with enhanced video editing comprehension abilities. Trained on VEU-50K training set, Oscars demonstrates superior performance across all VEU tasks, outperforming the leading open-source model LLaVA-Onevision~\cite{wang2024qwen2} by 28.3\% and surpassing Gemini by 4.0\%\cite{team2023gemini}, as illustrated in Figure~\ref{fig:rader_map} and Table~\ref{tab:main_benchmark}. Additionally, Oscars achieves improved performance on several general video understanding benchmarks~\cite{Video-MME, liu2024tempcompass, MVBench}, with average gains of 5.1\%, 4.8\%, and 6.6\% on reasoning-related tasks in Video-MME~\cite{Video-MME}, MVBench~\cite{MVBench}, and TempCompass~\cite{liu2024tempcompass}, demonstrating the value of video editing data in fostering high-level abstract learning, which in turn benefits general video understanding capabilities.

Our contributions are summarized as follows:
\begin{itemize}
    \item We introduce VEU-Bench, the first comprehensive benchmark designed to evaluate Vid-LLM performance on video editing understanding tasks, covering 19 tasks across 10 dimensions and 3 levels. We also provide 50k high-quality data points to advance research in video editing understanding.
    \item We conduct a thorough assessment of the current state-of-the-art Vid-LLMs on the proposed VEU-Bench. Our findings reveal that current Vid-LLMs struggle with understanding video editing components, especially in reasoning and judging tasks.
    \item By fine-tuning on carefully curated VEU-50k dataset, our proposed baseline Vid-LLM, Oscars, surpasses existing state-of-the-art models with a 28.3\% overall improvement on VEU-Bench and an average 3\% improvement on other general video benchmarks.
\end{itemize}

\section{Related Work}
\subsection{Video Editing Understanding}
Video editing involves complex tasks that go beyond the pixel-level manipulations of video frame~\cite{sun2024diffusion}. Editing video clips~\cite{wohl2002editing} is especially challenging in video understanding, requiring a nuanced grasp of both content and editing principles\cite{dancyger2018technique}. Prior studies have explored various technical aspects of editing, including camera shot settings~\cite{burch2014theory, argaw2022anatomy, vacchetti2022movie, huang2020movienet}, transitions between shots~\cite{shen2022autotransition}, visual effects~\cite{gu2024edit3k,xu2024beyond}, and cut types~\cite{pardo2021learning, pardo2022moviecuts, chen2023match}. Progress in video editing understanding has been hindered by the lack of large, high-quality public datasets, the subjective nature of editing quality assessment, and the limitations of current video understanding models.

\subsection{Video Large Language Model}
Recently, the advancement of Vid-LLMs has driven a significant leap forward in the video domain. This progress is highlighted by the introduction of pioneering Vid-LLMs~\cite{cheng2024videollama,fei2024video,zhang2024internlm,chen2024far,yao2024minicpm,wang2024qwen2,lin2024vila, zhang2024video,zhang2023video,lin2023video,ye2024mplug,chen2024sharegpt4video,huang2024vtimellm,maaz2023video, wu2024zero, li2024frame, liu2024omniclip, cao2024reframe}. By integrating a visual encoder and training the projector on large-scale multimodal instruction datasets, these models demonstrate remarkable performance in video understanding tasks.
VideoChatGPT~\cite{maaz2023video} and LLaVA-Video~\cite{zhang2024video} utilize LLM and VLM to generate video instruction-tuning data. Works like Qwen2-VL~\cite{wang2024qwen2}, VideoLLaMA2~\cite{cheng2024videollama}, Video-CCAM~\cite{fei2024video} and InternLM~\cite{zhang2024internlm} proposed new design insights for video-llm construction. While these models excel in general video understanding benchmarks, their potential applications in video editing tasks remain underexplored. EditQA-2k~\cite{xu2024beyond} explored the capability of Vid-LLMs in analyzing edited video content, but a comprehensive evaluation of Vid-LLMs in understanding video editing components remains absent. 

\subsection{Video Understanding Benchmark}
As Video Large Language Models (Vid-LLMs) advance, various benchmarks~\cite{Video-MME, MVBench, zhou2024mlvu, wu2024longvideobench, wang2024lvbench, liu2024tempcompass,patraucean2024perception, wu2024video} have been developed to evaluate their capabilities. Benchmarks like Video-MME~\cite{Video-MME}, MVBench~\cite{MVBench}, and TempCompass~\cite{liu2024tempcompass} focus on general understanding. Others, such as LongVideoBench~\cite{wu2024longvideobench}, MLVU~\cite{zhou2024mlvu}, and LVBench~\cite{wang2024lvbench}, examine long-form videos processing ability of Vid-LLM. While these benchmarks offer insights into the general understanding of Vid-LLMs, they overlook complex tasks like video editing understanding that require advanced reasoning. Additionally, most benchmarks are formatted with only multiple-choice QA, making it hard to evaluate the reasoning abilities that require evaluation in an open-ended way. Therefore, we propose VEU-Bench, a comprehensive video editing understanding benchmark to thoroughly evaluate Vid-LLMs’ video editing understanding ability with 50K high-quality video editing understanding data.

\section{Benchmark}
\subsection{Overview}
\begin{figure}[t]
    \centering
    \begin{subfigure}{0.23\textwidth}
        \centering
        \includegraphics[width=\textwidth]{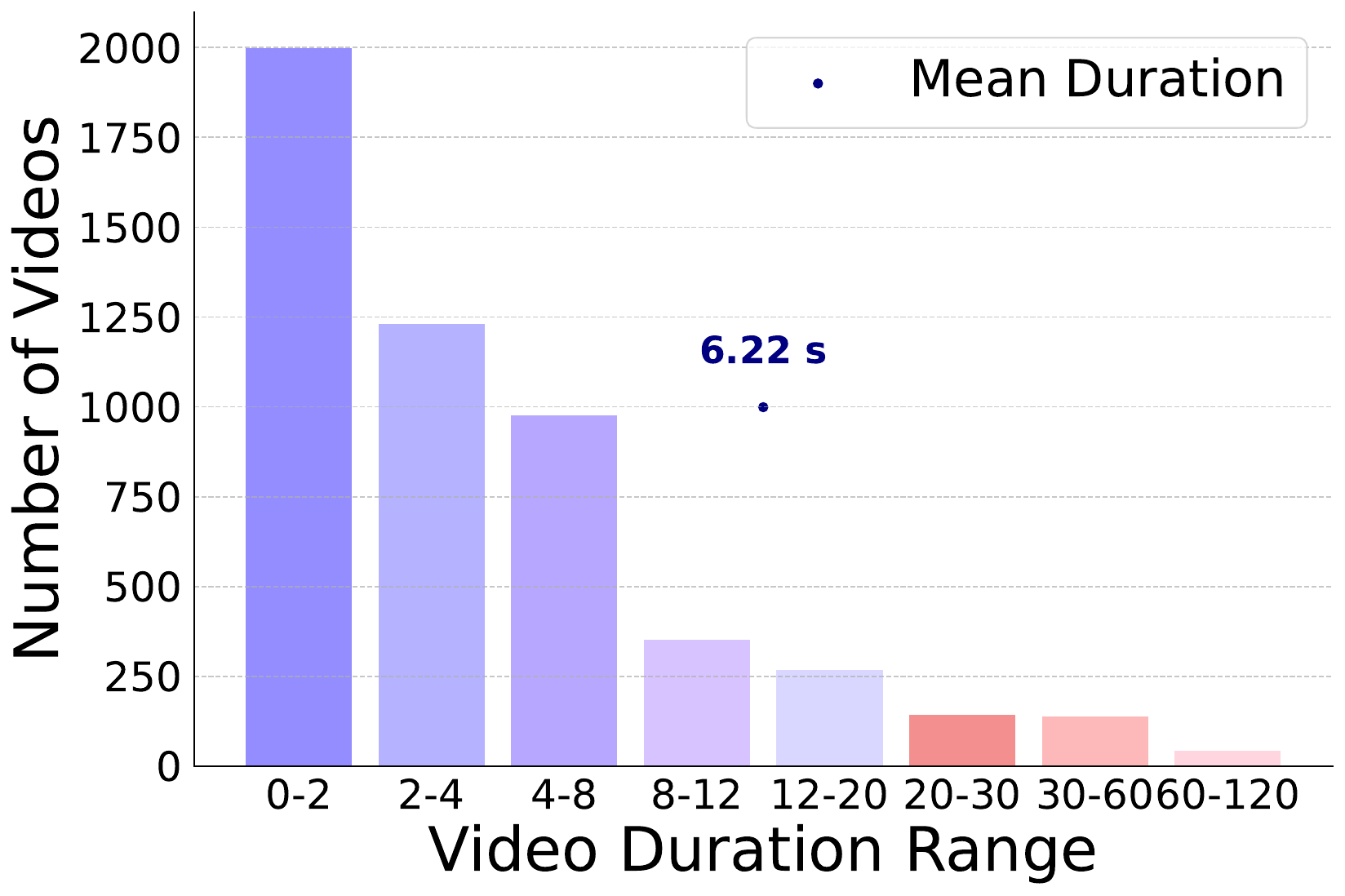}
        \label{fig:fig1}
    \end{subfigure}
    \begin{subfigure}{0.23\textwidth}
        \centering
        \includegraphics[width=\textwidth]{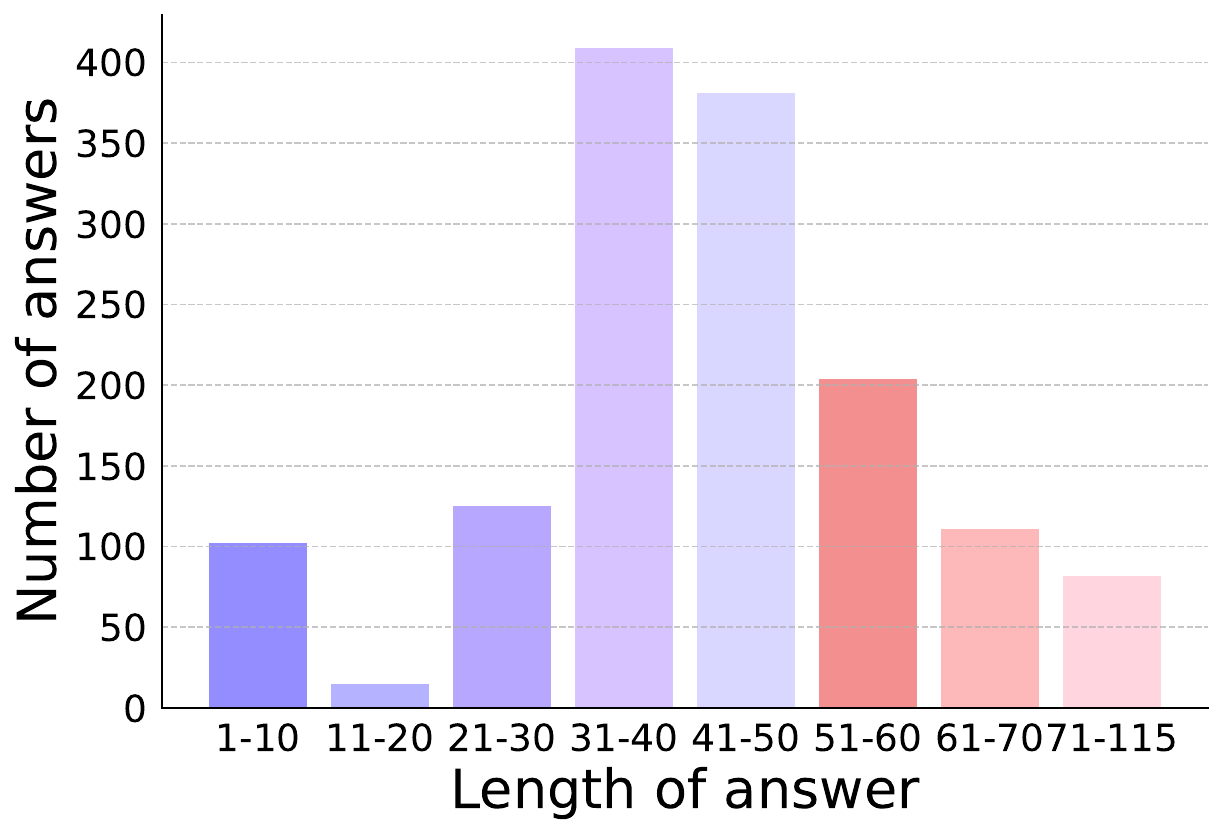}
        \label{fig:fig2}
    \end{subfigure}
    \begin{subfigure}{0.25\textwidth}
        \centering
        \includegraphics[width=\textwidth]{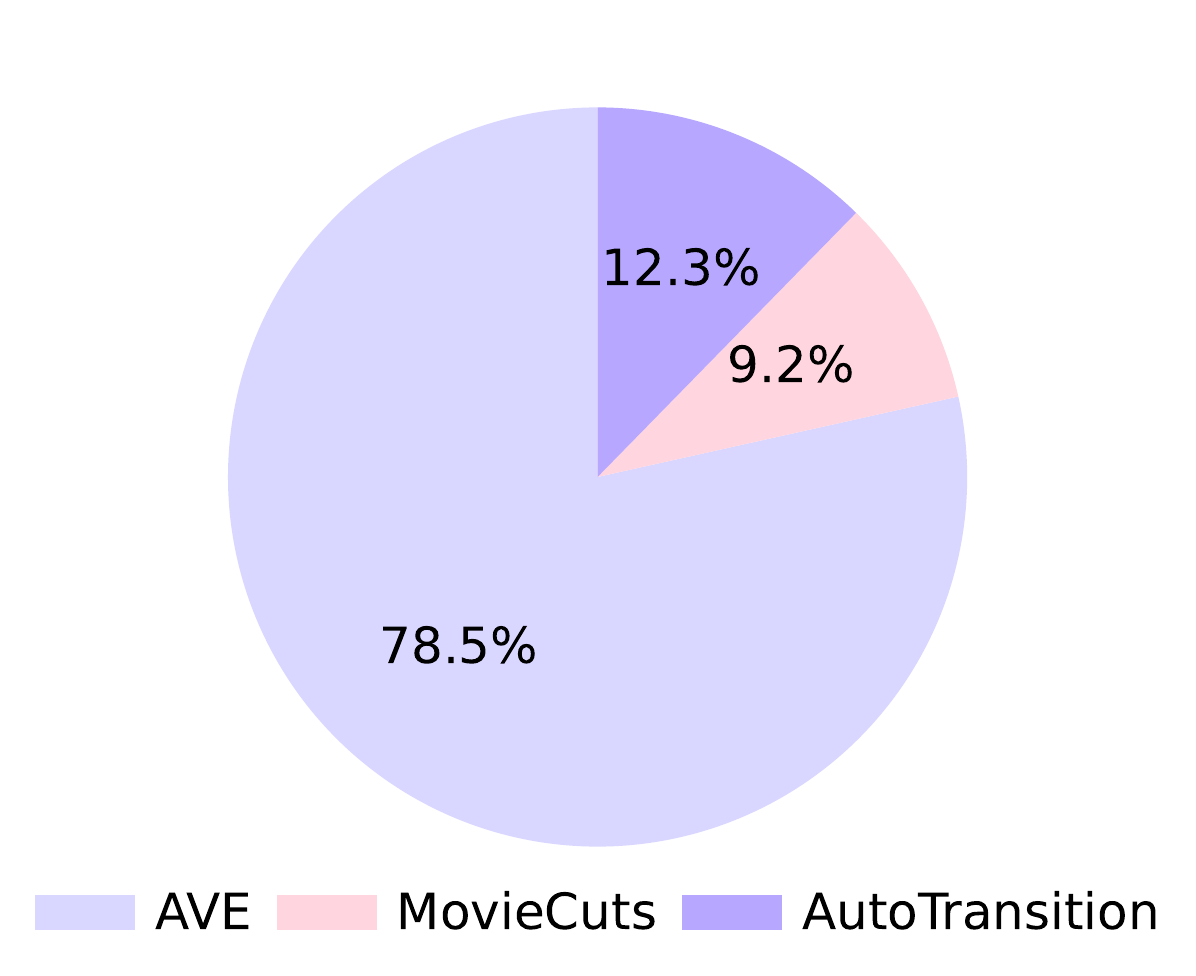}
        \label{fig:fig3}
    \end{subfigure}
    \begin{subfigure}{0.22\textwidth}
        \centering
        \includegraphics[width=\textwidth]{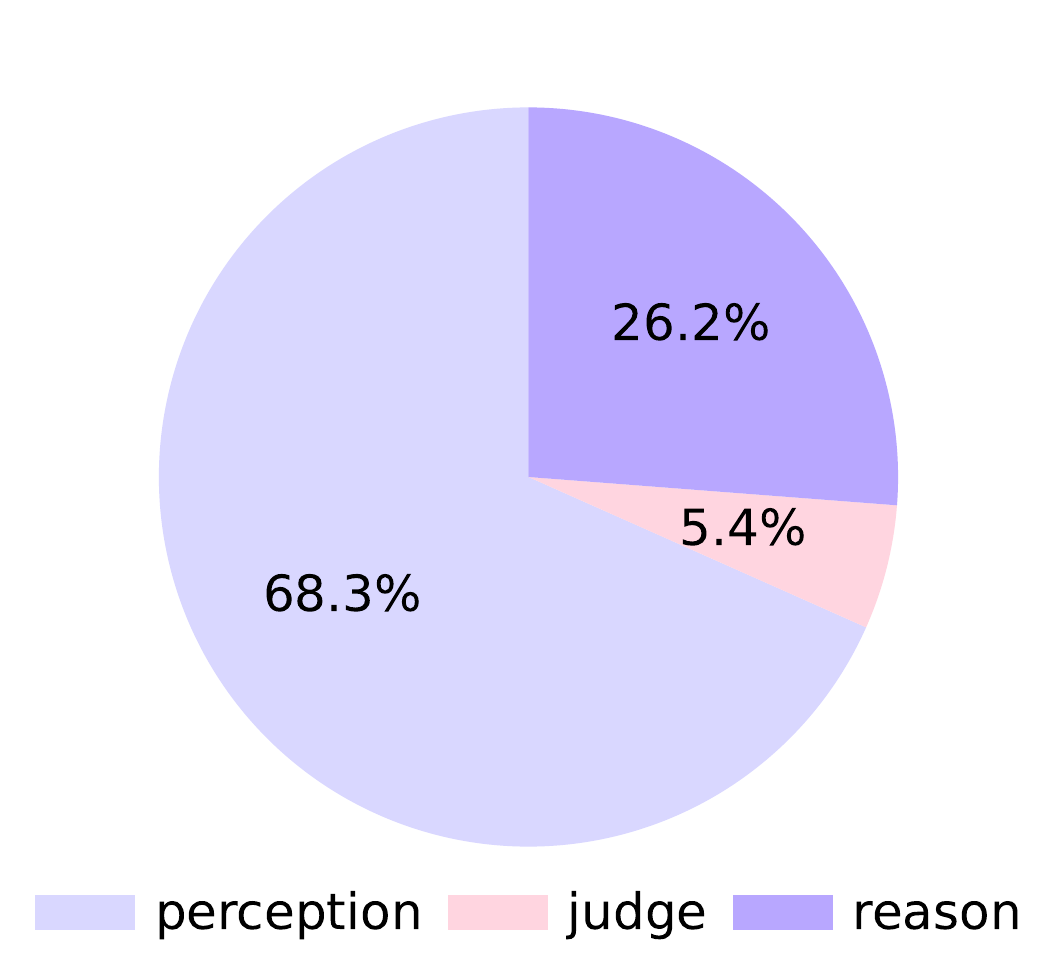}
        \label{fig:fig4}
    \end{subfigure}
    \caption{The static of our proposed VEU-Bench.}
    \label{fig:benchmark_overview}
    \vspace{-10pt}
\end{figure}

As shown in Figure~\ref{fig:benchmark_overview}, VEU-Bench encompasses editing components across 10 dimensions and 3 levels, including a total of 30,000 videos and 49,536 QA samples. The training set contains 45,154 samples, and the test set contains 4,382 samples. The video durations range from 1 to over 60 seconds, with the majority between 1 and 12 seconds. We compare our proposed VEU-Bench with previous benchmarks in Table~\ref{tab:dataset_comparison}. We begin by introducing the task definition of VEU-Bench in Section~\ref{sec:task_definition}, covering both task dimensions and levels. Next, the dataset construction process is detailed in Section~\ref{sec:dataset_construction}. Finally, we present the evaluation methods in Section~\ref{sec:dataset_evaluation}.

\begin{table}[t]
    \centering
    \caption{\textbf{The comparison between VEU-Bench and previous VEU benchmarks.} VEU-Bench encompasses a wider range of video editing components and includes high-level reasoning and judgment tasks.}
    \resizebox{\linewidth}{!}{
    \begin{tabular}{lc|ccc|c|c}
        \toprule
        \textbf{Dataset}                            & \textbf{Size} & \textbf{Shot} & \textbf{Cut} & \textbf{Transition} & \textbf{Reasoning} & \textbf{Raw Video}  \\
        \midrule
        MovieNet~\cite{huang2020movienet}            & 1k            & \Checkmark    & \XSolidBrush & \XSolidBrush        & \XSolidBrush       & \XSolidBrush  \\
        AVE~\cite{argaw2022anatomy}                  & 200k          & \Checkmark    & \XSolidBrush & \XSolidBrush        & \XSolidBrush       & \XSolidBrush  \\
        AutoTransition~\cite{shen2022autotransition} & 30k           & \XSolidBrush  & \XSolidBrush & \Checkmark          & \XSolidBrush       & \Checkmark  \\
        MovieCuts~\cite{pardo2022moviecuts}          & 174k          & \XSolidBrush  & \Checkmark   & \XSolidBrush        & \XSolidBrush       & \Checkmark  \\
        Edit3k~\cite{gu2024edit3k}                   & 3K            & \Checkmark    & \XSolidBrush & \Checkmark          & \XSolidBrush       & \XSolidBrush  \\
        EditedVideo2K~\cite{xu2024beyond}            & 2K            & \XSolidBrush  & \XSolidBrush & \Checkmark          & \Checkmark         & \Checkmark  \\
        \midrule
        Oscar (ours)                                & 50K           & \Checkmark    & \Checkmark   & \Checkmark          & \Checkmark         & \Checkmark  \\
        \bottomrule
    \end{tabular}}
    \label{tab:dataset_comparison}
\end{table}

\begin{figure*}[t]
    \centering
    \includegraphics[width=0.9\textwidth]{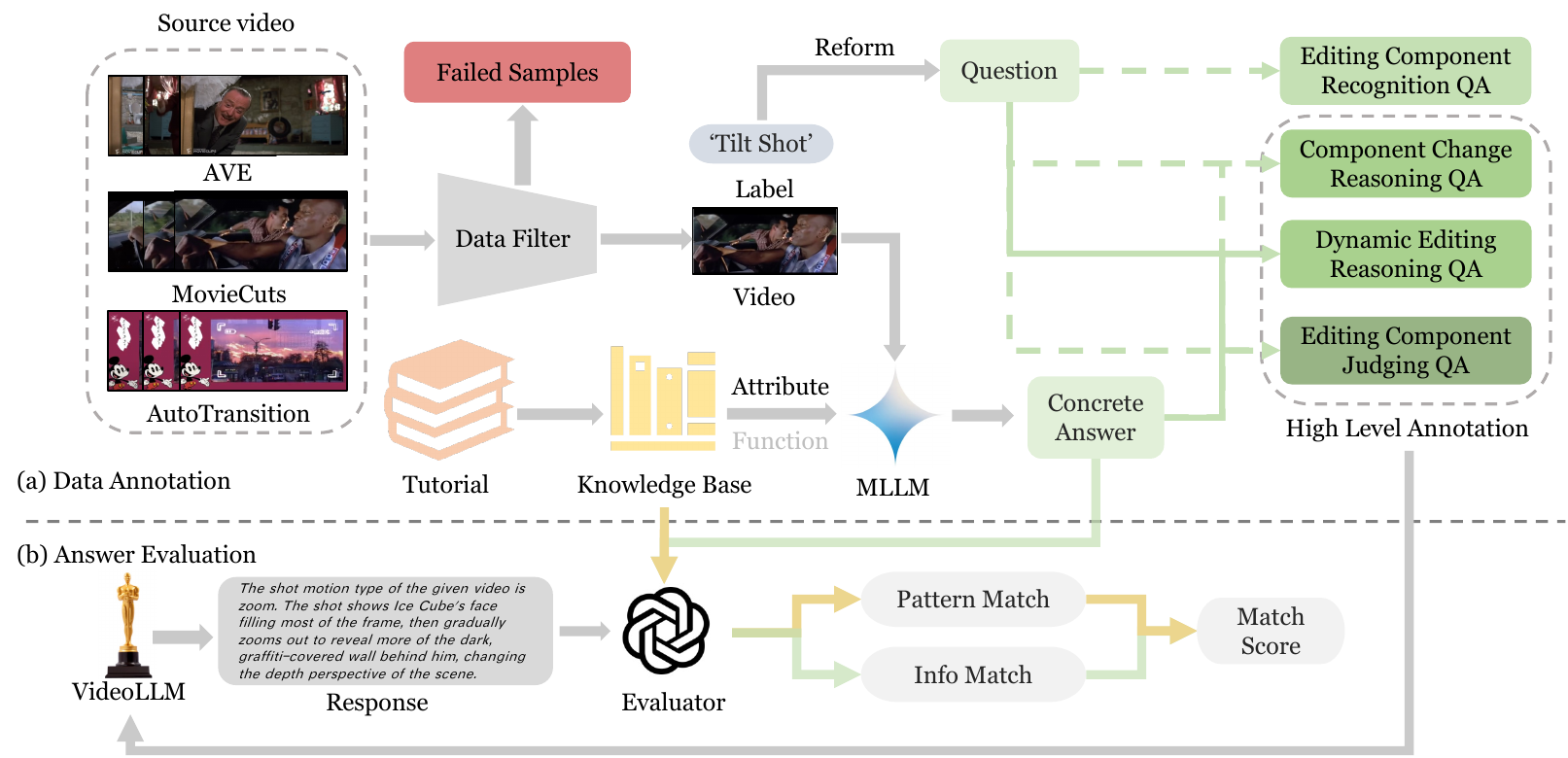}
    \caption{\textbf{The overview of our data annotation pipeline.} (a) shows the data annotation process for reasoning and judging tasks. Based on an established knowledge base, the annotator selects the most relevant attribute or function and reformulates a video-specific answer to create the QA pair. (b) Indicate evaluation mechanism, the response is matched against the corresponding abstract feature in the knowledge base, as well as compared with the annotated answer to calculate an overall score.}
    \label{fig:dataset_pipeline}
\end{figure*}

\subsection{Task Definition}
\label{sec:task_definition}
In this section, we provide detailed task definition of VEU-Bench, focusing on two aspects: task dimensions and task levels. Detailed descriptions are attached in the Appendix.
\subsubsection{Task Dimensions}
Referring to industry definitions~\cite{wohl2002editing, dancyger2018technique}, we classify video editing components into three categories: \textit{Intra-frame}: Involves only a single frame. \textit{Intra-shot}: Spans multiple frames within a single scene. \textit{Inter-shot}: Covers multiple shots across different scenes. Each category includes various editing components, resulting in a total of 10 distinct dimensions as follows:

\noindent \textbf{Intra-frame.} (1) \textit{Shot Size:} The proportion of the setting or subject that is visible within the frame of a shot. (2) \textit{Shot Angle:} The angle or perspective from which the camera views the subject, influencing how the subject appears in the frame (e.g., high-angle, low-angle). (3) \textit{Shot Location:} The physical environment or setting in which the scene takes place, providing context or background. (4) \textit{Shot Subject:} The main subject that is highlighted or conveyed within the shot. (5) \textit{Shot Type:} The composition of a shot with respect to the number of featured subjects and their physical relationship to each other as well as to the camera. (6) \textit{Shot color:} The color grading of video clips to create a specific visual tone or mood.

\noindent \textbf{Intra-shot.} (7) \textit{Shot motion:} The movement of the camera during the process of taking a shot. (8) \textit{Shot speed:} The playback speed to create different effects within a shot. 

\noindent \textbf{Inter-shot.} (9) \textit{Cut:} Composed of two adjacent shots and transition between them, without special visual effects and achieved simply by cutting. (10) \textit{Transition:} Moving from one shot to next with special visual effects, creating either an intentional visual impact as one clip changes to another.

\subsubsection{Task Levels} 
While previous VEU benchmarks~\cite{shen2022autotransition, pardo2022moviecuts, argaw2022anatomy} mainly focus on recognition, we expand the scope of the task to include reasoning and judging, thereby enabling a more comprehensive evaluation. The task levels are constructed as follows: 

\noindent \textbf{Recognition.} At this level, the model should correctly classify different categories within various task dimensions, in form of multichoice question answering.

\noindent \textbf{Reasoning.} At this level, the model is required not only to correctly identify corresponding elements but also to provide evidence and rationale. Two aspects are included in the reasoning task. First, the reasoning of dynamic editing elements e.g. cut and transition. Second, reasoning on changing static editing elements e.g. change of shot size and shot angle.  For these tasks, the model must spot and explain changes within a single shot.

\noindent \textbf{Judging.} At this level, model is expected to evaluate the functions and benefits of editing elements, like cut and shot types, by interpreting their role within video based on both the content and the creator’s intent. Judging assesses model’s ability to applying its understanding to explain how each editing choice contributes to storytelling or desired impact within a real editing context.

\subsection{Dataset Construction}
\label{sec:dataset_construction}
\subsubsection{Video Collection}
We begin by collecting video data from existing video editing datasets: AVE~\cite{argaw2022anatomy}, MovieCuts~\cite{pardo2022moviecuts}, and AutoTransition~\cite{shen2022autotransition}. We focus on short clips to enable Vid-LLMs to process fundamental units effectively and ensure compatibility within context length limitations~\cite{chen2024far}. We exclude unsuitable dimensions, filter out incorrect annotations using Gemini-1.5-pro~\cite{team2023gemini}, and balance data distribution to mitigate the long-tail effect from dominant labels.

\subsubsection{Automatic Annotation} 
\noindent \textbf{Recognition.} In this task, we introduce two new dimensions: shot color and shot speed. Shot color labels are generated based on HSV color thresholds~\cite{kim2020automatic}, while shot at different speeds is created using ffmpeg~\cite{tomar2006converting} to apply speed adjustments, including both fast and slow motion effects. Due to broad variety of transition types, transition recognition task randomly samples three other categories per sample to structure multiple-choice questions. For other recognition tasks, we include all dimensions in question options.

\noindent \textbf{Reasoning \& Judging.} Reasoning and judging video editing components require costly manual annotation due to the need for expert knowledge. Direct annotation by MLLM also proves challenging, as evidenced in Figure~\ref{fig:rader_map}, even top-performing models like Gemini\cite{team2023gemini} struggle with VEU tasks. To overcome this hindrance, we propose an automatic pipeline that leverages the ontology-based system\cite{khurana2013study} grounded in an abstract knowledge base. As shown in Figure~\ref{fig:dataset_pipeline}, for each editing component, we construct a professional knowledge base by integrating definitions from existing video editing tutorial~\cite{wohl2002editing}. Specially, for reasoning tasks, we define “key attributes” describing abstract patterns of each dimension. Each dimension contains features that serve as accordance for reasoning and identification. For the Judging task, we define the functions of editing elements within video content, ensuring video-independent applicability. During annotation, MLLM selects the most relevant attributes or functions based on the video content, and then abstract terms like “object” or “scene” in pre-defined features are replaced with specific terms to ensure content relevance. For example, attributes for \textit{“match cut”} include elements like \textit{“connecting two similarly shaped objects across frames.”} While annotating a particular video, this feature will be rewritten as \textit{“a match cut connects a similarly shaped bone and spaceship across frames.”} This annotation process simplifies open-ended reasoning to a rewriting task, enables our benchmark to deliver high-quality video editing annotations with minimal human involvement and guarantees LLM annotators to generate well-crafted responses with less need for deep video editing knowledge. After completing the construction of the dataset, we conducted manual reviews and user studies to further validate the data quality. Details can be found in the appendix.

\begin{table*}[t]
\centering
\caption{\textbf{The performance of current state-of-the-art Vid-LLMs and the proposed Oscars on VEU-Bench.} The best performance is highlighted in bold and the second-best is underlined. Notably, Oscars surpasses the base model Qwen2-VL-7B by 39.6\% and outperforms Gemini-1.5-pro by 4\%, achieving performance comparable to GPT-4o.}
\resizebox{\textwidth}{!}{
\begin{tabular}{c|c|ll|llllllll|ll}
\toprule
\textbf{Level} & \textbf{Dimension} & \rotatebox{70}{Gemini-1.5-Pro~\cite{team2023gemini}} & \rotatebox{70}{GPT-4o~\cite{gpt4o}} & \rotatebox{70}{InternLM-X-7B~\cite{zhang2024internlm}} & \rotatebox{70}{InternVL2-8B~\cite{chen2024far}} & \rotatebox{70}{LLaVA-OV-7B~\cite{li2024llava}} & \rotatebox{70}{LLaVA-Video-7B~\cite{zhang2024video}} & \rotatebox{70}{Video-CCAM-7B~\cite{fei2024video}} & \rotatebox{70}{VideoLLaMA2-7B~\cite{cheng2024videollama}} & \rotatebox{70}{VILA-8B~\cite{xue2024longvila}} & \rotatebox{70}{Oryx-7B~\cite{liu2024oryx}} & \rotatebox{70}{Qwen2-VL-7B~\cite{wang2024qwen2}} & \rotatebox{70}{\textbf{Oscars-7B(ours)}} \\
\midrule
\multicolumn{2}{c|}{Frame Numbers} & 1FPS          & 8             & 64   & 16   & 16   & 16   & 96   & 32   & 16   & 32    & 1FPS  & 1FPS \\
\midrule
\multirow{10}{*}{{Recognition}} & Shot Subject  & 77.3          & \uline{79.8}  & 52.2 & 71.5 & 76.8 & 52.2 & 53.3 & 52.5 & 78.7 & 55.8  & 71.3 & \textbf{80.4} (\textcolor{LimeGreen}{+9.1})\\
~                               & Shot Color    & 49.3          & 49.1 & 48.0 & 44.7 & 39.3 & 48.0 & 38.0 & 37.3 & 29.3 & \textbf{53.3}  & 45.3 & \uline{51.3} (\textcolor{LimeGreen}{+6.0})\\
~                               & Shot Size     & 55.6          & \uline{63.3}  & 53.0 & 51.5 & 22.4 & 43.3 & 52.0 & 50.7 & 49.3 & 51.5  & 51.7 & \textbf{66.2} (\textcolor{LimeGreen}{+14.5})\\
~                               & Shot Angle    & \uline{54.5}  & \textbf{60.2} & 36.0 & 36.7 & 39.8 & 36.0 & 39.2 & 33.4 & 39.2 & 36.8  & 45.8 & 48.6 (\textcolor{LimeGreen}{+2.8}) \\
~                               & Shot Location & \textbf{90.2} & 89.1          & 88.4 & 89.5 & 84.9 & 88.4 & 87.7 & 84.2 & 87.7 & 76.5  & 86.7 & \uline{90.0} (\textcolor{LimeGreen}{+3.3}) \\
~                               & Shot Type     & \uline{76.4}  & \textbf{83.0} & 66.4 & 72.7 & 71.1 & 66.4 & 59.2 & 55.2 & 62.7 & 60.6  & 62.2 & 73.7 (\textcolor{LimeGreen}{+11.5}) \\
~                               & Shot Motion   & \textbf{36.3} & \uline{35.2}  & 30.2 & 22.9 & 20.6 & 30.2 & 24.0 & 21.4 & 18.1 & 19.5  & 27.0 & 34.1 (\textcolor{LimeGreen}{+7.1}) \\
~                               & Shot Speed    & \uline{35.1}  & 33.3          & 33.3 & 33.3 & 32.0 & 33.3 & 33.3 & 26.6 & 33.3 & 33.3  & 36.0 & \textbf{40.1} (\textcolor{LimeGreen}{+6.8}) \\
~                               & Transition    & 35.3          & \uline{51.1}  & 25.4 & 15.1 & 49.8 & 25.3 & 13.0 & 38.9 & 40.5 & 26.6  & 27.5 & \textbf{55.9} (\textcolor{LimeGreen}{+29.7}) \\
~                               & Cut Type      & \uline{32.1}  & \textbf{42.4} & 17.1 & 21.4 & 17.6 & 17.1 & 16.7 & 17.1 & 17.6 & 17.6  & 16.2 & 28.7 (\textcolor{LimeGreen}{+12.5}) \\
\midrule
\rowcolor{Gray!30}\multicolumn{2}{c|}{\textbf{Score$_{mc}$}} & 2.71 & \textbf{2.93} & 2.25 & 2.29 & 2.27 & 2.20 & 2.08 & 2.09 & 2.28 & 2.20 & 2.33 & \uline{2.85} (\textcolor{LimeGreen}{+0.52}) \\
\midrule
\multirow{7}{*}{{Reasoning}} & Shot Size     & 1.05         & 1.28 & 0.88 & 2.52 & 0.86 & 0.54 & \uline{1.55} & \textbf{1.84} & 0.97 & 0.55 & 0.80 & 1.13 (\textcolor{LimeGreen}{+0.16}) \\
~                            & Shot Angle    & \uline{1.30} & 1.03          & 0.90 & 0.85 & 0.71 & 0.72 & 0.70 & 0.70 & 0.08 & 0.60 & 0.24 & \textbf{2.34} (\textcolor{LimeGreen}{+2.10}) \\
~                            & Shot Location & \textbf{3.96}& \uline{3.91}  & 3.63 & 3.27 & 3.64 & 1.11 & 3.34 & 2.12 & 2.02 & 2.78 & 2.73 & 3.53 (\textcolor{LimeGreen}{+1.50})\\
~                            & Shot Type     & 3.46         & \textbf{3.85} & 2.84 & 2.90 & 3.00 & 0.66 & 1.70 & 1.33 & 2.60 & 2.35 & 2.61 & \uline{3.81} (\textcolor{LimeGreen}{+1.07}) \\
~                            & Shot Motion   & 0.99         & \uline{1.78}  & 0.68 & 0.19 & 1.15 & 0.67 & 1.29 & 1.20 & 1.15 & 1.12 & 1.29 & \textbf{1.92} (\textcolor{LimeGreen}{+0.77}) \\
~                            & Transition    & \textbf{1.26}& 0.77          & 0.19 & 0.24 & 0.72 & 0.11 & 0.35 & 0.35 & 0.30 & 0.14 & 0.21 & \uline{0.91} (\textcolor{LimeGreen}{+0.70}) \\
~                            & Cut Type      & \uline{1.94} & \textbf{2.25} & 0.78 & 1.16 & 1.05 & 1.43 & 0.75 & 0.81 & 0.87 & 0.70 & 0.80 & 1.58 (\textcolor{LimeGreen}{+0.78}) \\
\midrule
\multirow{2}{*}{Judging}  & Shot Type     & 3.65 & \textbf{3.95} & 3.04 & 2.97 & 3.15 & 0.96 & 1.42 & 1.65 & 2.36 & 2.08 & 2.66 & \uline{3.74} (\textcolor{LimeGreen}{+1.08}) \\
~                            & Cut Type      & 1.98 & \textbf{2.39} & 0.34 & 0.19 & 0.94 & 0.97 & 0.32 & 0.91 & 0.51 & 0.68 & 0.41 & \uline{2.04} (\textcolor{LimeGreen}{+1.63}) \\
\midrule
\rowcolor{Gray!30}\multicolumn{2}{c|}{\textbf{Score$_{oe}$}} & 2.11 & \uline{2.36} & 1.48 & 1.59 & 1.69 & 0.80 & 1.27 & 1.21 & 1.21 & 1.22 & 1.31 & \textbf{2.23} (\textcolor{LimeGreen}{+1.17})\\
\midrule
\rowcolor{Gray!30}\multicolumn{2}{c|}{\textbf{Score$_{all}$}}   & 2.44 & \textbf{2.64} & 1.79 & 1.94 & 1.98 & 1.50 & 1.68 & 1.65 & 1.75 & 1.71 & 1.82 & \uline{2.54} (\textcolor{LimeGreen}{+0.72}) \\
\bottomrule
\end{tabular}
\label{tab:main_benchmark}
\vspace{-10pt}
}
\end{table*}

\subsection{Evaluation}
\label{sec:dataset_evaluation}
\subsubsection{Metrics}
For the Reasoning and Judging tasks, which require open-ended answers, we adopt an approach inspired by Video-ChatGPT~\cite{Maaz2023VideoChatGPT}. We leverage GPT-4~\cite{gpt4o} to assign relative scores to generated predictions. Directly scoring responses against correct answers, however, may lead to inflated scores, as responses often contain correct descriptive information but misinterpret editing patterns. To address this, we propose a scoring system that incorporates pattern-matching regularization based on a knowledge base constructed for dataset annotation, as shown in Figure~\ref{fig:dataset_pipeline} (b).

Specifically, given an editing component ontology $O$ (attributes for reasoning tasks and functions for judging tasks), ground-truth answer$A_{\text{gt}}$, and predicted answer $A_{\text{pd}}$, matching score between response and answer is defined as:
$$
S_{\text{match}} = \frac{PM(A_{\text{pd}}, O) + IM(A_{\text{pd}}, A_{\text{gt}})}{2} \in[0,5]
$$
where $PM$ (pattern matching score) measures how well $A_{\text{pd}}$ aligns with the editing pattern ontology $O$, focusing on the video-agnostic editing pattern. $IM$ (information matching score) assesses the alignment of specific objects and visual details between $A_{\text{pd}}$ and $A_{\text{gt}}$. Additionally, the accuracy $Acc$ of the editing component is computed by the LLM for open-ended questions. To evaluate performance on reasoning and judging tasks, we scale accuracy to a 0-5 range and combine it with the matching score, resulting in:
$$
S_{oe} = \frac{(5 \times Acc + S_{\text{match}})}{2} \in[0,5]
$$

For Recognition tasks, we use accuracy as the metric, scaling it to the 0-5 range in sync with the reasoning and judging metrics, defined as $S_{mc} = 5 \times Acc \in[0,5]$.

\subsubsection{Prompt Set} 
Video editing understanding requires domain-specific knowledge,  therefore, a simple prompt that directly queries the model cannot thoroughly evaluate the VEU ability of Vid-LLMs. Consequently, we design two extra prompts to optimize model performance to achieve the best results. First, we incorporate \textbf{context prompt} and integrate detailed definitions of editing components into the prompt to help the model better grasp the concept of these components and focus on understanding the video content. Second, \textbf{guidance prompt} that underscores general video task requirements are included. Also, we adapt different guidance prompts for each model while maintaining prompt semantics. Prompts samples are included in the Appendix.

\section{Experiments}
\subsection{Implementation Details}
We conduct extensive benchmarking across mainstream Vid-LLMs, including open-source models LLaVA-Video~\cite{zhang2024video}, InternLM-X~\cite{zhang2024internlm}, InternVL~\cite{chen2024far}, LLaVA-OneVision~\cite{li2024llava}, Qwen2-VL~\cite{wang2024qwen2}, Video-CCAM~\cite{fei2024video}, VideoLLaMA2~\cite{cheng2024videollama}, Oryx~\cite{liu2024oryx}, ViLA~\cite{lin2024vila} and proprietary models include Gemini-1.5-Pro~\cite{team2023gemini} and GPT-4o~\cite{gpt4o}.

For model training, we use Qwen2-VL-7B~\cite{wang2024qwen2} as the base model and apply LoRA~\cite{hu2021lora} fine-tuning with $r=16$ and $\alpha=32$. The learning rate is set to $1e-4$, weight decay to $0.01$, and warmup ratio to $0.05$. The model is optimized using AdamW~\cite{loshchilov2017decoupled}. For each video, frames are sampled at 1 fps with a maximum limit of 64 frames. All experiments are conducted on 4 A100 GPUs.

\subsection{Main Results}
\subsubsection{Quantitative Results}
We evaluate VEU-Bench on 11 state-of-the-art Vid-LLMs, as shown in Table~\ref{tab:main_benchmark}. Current Vid-LLMs exhibit poor performance across all benchmark dimensions, indicating a lack of capability in recognizing and reasoning about video editing components.

In the recognition task, Vid-LLMs perform significantly better at identifying intra-frame editing components than inter-frame and inter-shot content. This is likely due to the simplicity of analyzing a single frame when recognizing intra-frame elements, which does not involve dynamic content changes. Furthermore, in the areas of shot speed and shot motion, models like VideoLLaMA2 perform even worse than random guessing. This is attributed to the use of a uniform sampling strategy, which limits their ability to effectively perceive the concept of speed. 

Current Vid-LLMs also show limited performance in reasoning and judging tasks, with an average score below 2 out of 5. Compared to the recognition level, 
involving changes makes intra-frame features harder to recognize. Additionally, they score lower on cut and transition aspects compared to recognition tasks. Most open-source models exhibit lower performance on judging tasks compared to reasoning tasks in the cut-type dimension. This may be attributed to the challenge of establishing connections between the function of editing elements and specific video content, which goes beyond simple pattern recognition.

In contrast, by training on VEU-50k, our expert model Oscars exhibits improvements across all dimensions compared to Qwen2-VL\cite{wang2024qwen2}, with gains of 22.3\% in Score\(_{mc}\) and 89.3\% in Score\(_{oe}\). Meanwhile, Oscars gains 28.3\% higher performance in Score\(_{all}\) compared to SOTA open-source model LLaVA-OneVision\cite{li2024llava}. It also achieves performance comparable to state-of-the-art commercial models GPT-4o and surpasses Gemini-1.5-pro with 4\%. In the more challenging dimensions cut and transition, Oscars demonstrates significantly better performance than the open-source Vid-LLM state-of-the-art, with improvements of 12.2\% and 10.9\% on Score\(_{mc}\) and Score\(_{oe}\) respectively, showing exceeding video editing understanding ability. For a more detailed discussion on the dimension-wise performance of Vid-LLMs, please refer to the Appendix.

\subsubsection{Qualitative Results}
\begin{figure*}[t]
    \centering
    \includegraphics[width=\textwidth]{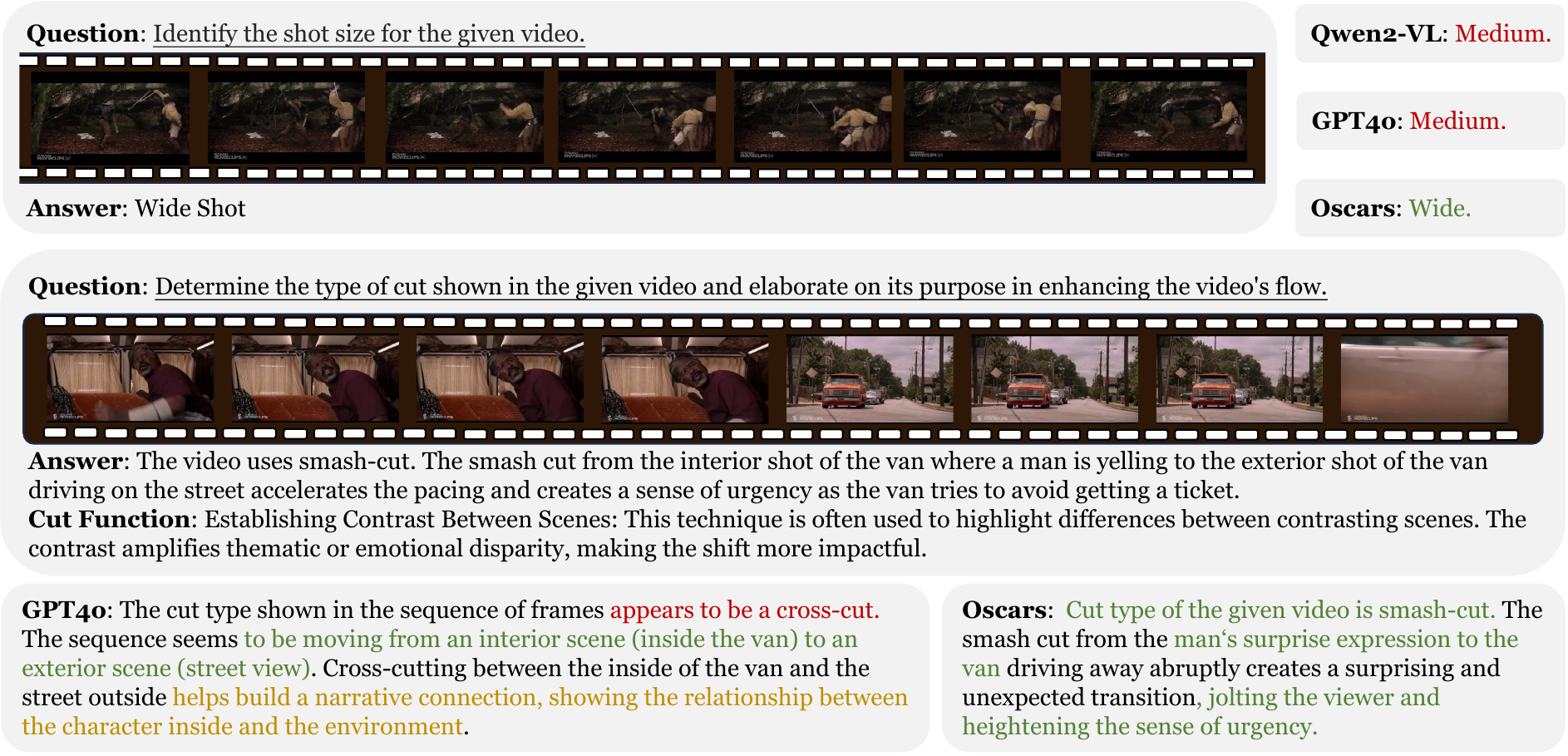}
    \caption{\textbf{The qualitative results of Oscars, GPT-4o and Qwen2-VL-7B on VEU-bench.} Green text indicates the correctly answered part; Red text indicates wrong information; Brown text indicates non-relevant judging on the effect of cut type.}
    \label{fig:Qualitative_Result}
    \vspace{-10pt}
\end{figure*}

We present a qualitative comparison among our model, Oscars, Qwen2-VL\cite{wang2024qwen2} and GPT-4o\cite{gpt4o} in Figure~\ref{fig:Qualitative_Result}. Oscars demonstrate superior video editing understanding, accurately identifying the \textit{``Wide''} shot size and function of the\textit{``Smash cut''} cut type. In contrast, Qwen2-vl and GPT-4o confused the nuanced difference between Medium and Wide shot size, and GPT-4o merely recognizes the content visible in the video and lacks the ability to reasonably infer the functionality of editing components. Additional visualization examples can be found in the Appendix.

\begin{table}[t]
    \centering
    \caption{\textbf{The performance of Oscars on representative dimension of general video understanding benchmarks.} Green numbers indicate improvement of Oscars compared to Qwen2-VL-7B.}
    \resizebox{\linewidth}{!}{
    \begin{tabular}{l|l|cc}
        \toprule
        Benchmark & Dimension & Qwen2-VL-7B & Oscars \\
        \midrule
        \multirow{3}{*}{VideoMME$_{short}$~\cite{Video-MME}}     & Attribute Perception & 73.0 & 80.3(\textcolor{LimeGreen}{+7.3}) \\
                                                             ~   & Action Reasoning     & 72.3 & 76.6(\textcolor{LimeGreen}{+4.3})   \\
                                                             ~   & Information Synopsis & 82.9 & 86.6(\textcolor{LimeGreen}{+3.7})  \\
        \midrule
        \multirow{3}{*}{MVBench~\cite{MVBench}}                  & Unexpected Action    & 72.0 & 78.0 (\textcolor{LimeGreen}{+6.0}) \\
                                                              ~  & State Change         & 47.0 & 52.5 (\textcolor{LimeGreen}{+5.5}) \\
                                                              ~  & Moving Direction     & 43.0 & 46.0 (\textcolor{LimeGreen}{+3.0}) \\
        \midrule
        \multirow{3}{*}{TempCompass~\cite{liu2024tempcompass}}   & Order               & 54.1 & 62.6 (\textcolor{LimeGreen}{+8.5}) \\
                                                              ~  & Direction           & 39.7 & 46.7 (\textcolor{LimeGreen}{+7.0})  \\
                                                              ~  & Speed               & 43.4 & 47.6 (\textcolor{LimeGreen}{+4.2}) \\
        \bottomrule
    \end{tabular}}
    \label{table:general_benchmark}
\end{table}

\begin{figure}[t]
    \centering
    \includegraphics[width=\linewidth]{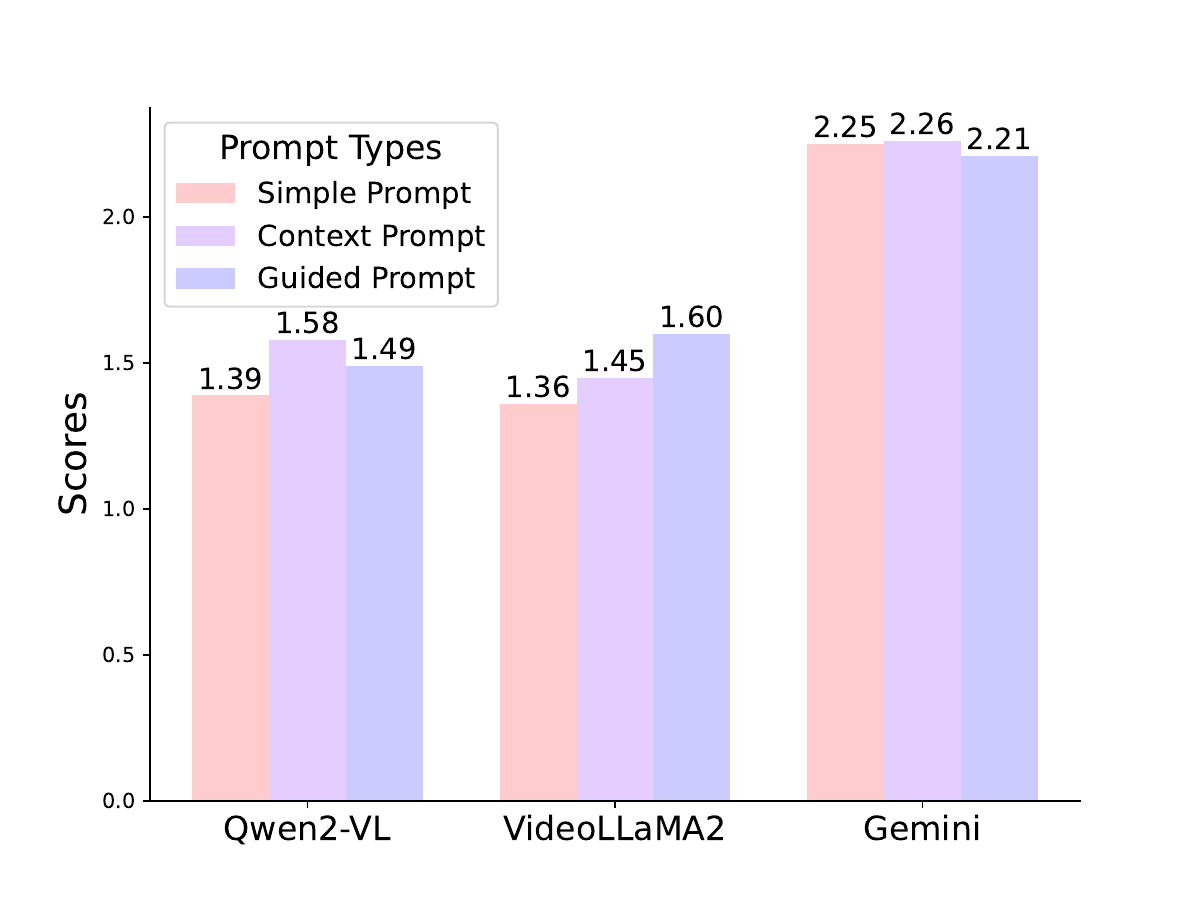}
    \caption{\textbf{Ablation results of prompt designs.}We conduct experiments on Qwen2-VL-7B, VideoLLaMA2-7B and Gemini-1.5-Pro.}
    \label{fig:prompt_ablation}
    \vspace{-10pt}
\end{figure}

\begin{table}[t]
    \centering
    \vspace{-10pt}
    \caption{\textbf{The concept experiment on Vid-LLMs.}}
    \resizebox{\linewidth}{!}{
    \begin{tabular}{lccccc}
        \toprule
        \textbf{Model}                        & \textbf{Intra-Frame} & \textbf{Intra-Shot} & \textbf{Inter-Shot} \\
        \midrule
        Gemini-1.5-Pro~\cite{team2023gemini}          & 2.76($\pm0.006$)     & 2.82($\pm0.006$)    & 2.82($\pm0.002$) \\
        Qwen2-VL~\cite{wang2024qwen2}         & 2.63($\pm0.010$)     & 2.62($\pm0.009$)    & 2.74($\pm0.005$) \\
        VideoLLaMA2~\cite{cheng2024videollama} & 2.52($\pm0.021$)     & 2.54($\pm0.014$)    & 2.57($\pm0.009$) \\
        LLaVA-Video~\cite{xue2024longvila}    & 2.65($\pm0.015$)     & 2.56($\pm0.012$)    & 2.62($\pm0.015$) \\
        \bottomrule
    \end{tabular}}
    \label{tab:concept_experiment}
    \vspace{-15pt}
\end{table}

\begin{figure}[t]
    \centering
    \includegraphics[width=\linewidth]{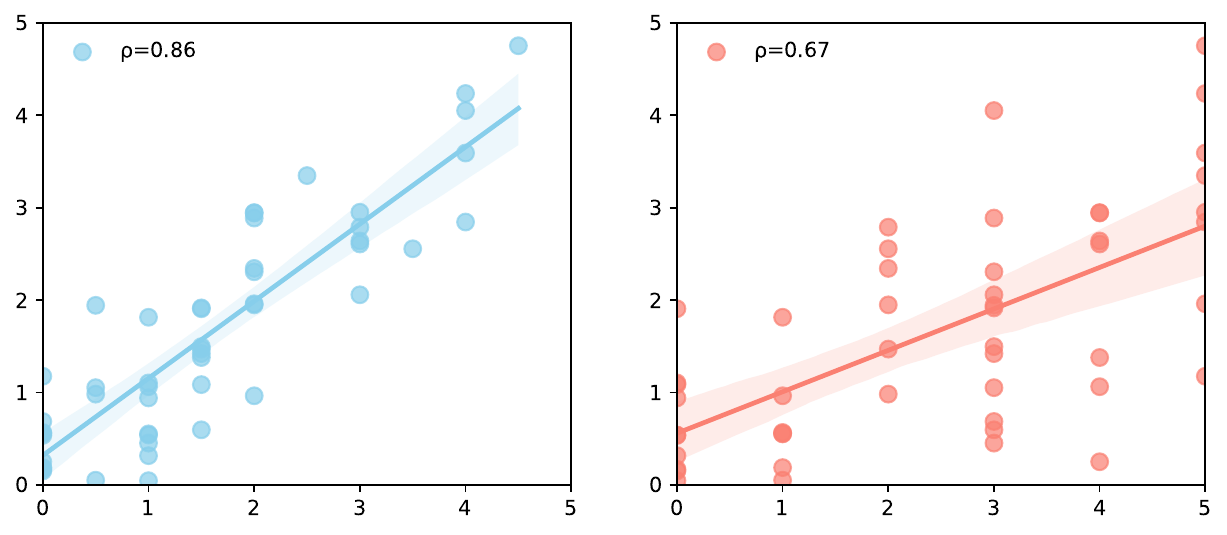}
    \caption{\textbf{The alignment between LLM (vertical axis) and human scores (horizontal axis).} Left figure indicates scoring with both IM and PM and right figure indicates scoring with only IM.}
    \vspace{-15pt}
    \label{fig:score alignment}
\end{figure}

\subsection{Deep Analysis}
\subsubsection{Results on General Video Benchmarks}
We evaluate the performance of Oscars on general video understanding benchmarks, as shown in Table~\ref{table:general_benchmark}. The results indicate that only by fine-tuning on 50k VEU-Bench data, the general video content understanding ability of the model is significantly enhanced. Specifically, Oscars achieves 7.3\% improvement on the attribute perception task in VideoMME-short\cite{Video-MME}, 5.5\% improvement on the state change task in MVBench\cite{MVBench} and 8.5\% improvement on the Ordering task in TempCompass\cite{liu2024tempcompass}. Given that VEU-50k contains domain-specific data with low similarity to general video understanding tasks, we attribute this improvement to the rich dynamic content and comprehension demands in the VEU data, which strengthen the temporal understanding ability of Vid-LLMs, leading to better performance on general tasks. We include full result on general video benchmarks in appendix

\subsubsection{Different Prompt Designs}
We explore the effects of three types prompt designs as discussed in Section~\ref{sec:dataset_evaluation}: simple prompt, context prompt, and guided prompt. As shown in Figure~\ref{fig:prompt_ablation}, compared to simple prompt, the context prompt helps activate editing-related knowledge in LLMs, enhancing the model’s ability to understand video editing elements. This leads to 6.6\% improvement on VideoLLaMA2~\cite{cheng2024videollama} and 13.7\% improvement on Qwen2-VL~\cite{wang2024qwen2}. Due to strong video understanding capabilities of Gemini-1.5-pro\cite{team2023gemini}, the effect of context prompt is less pronounced. Guidance prompts are more beneficial for models with weaker general video understanding capabilities, providing a 10\% improvement for VideoLLaMA. For Qwen2-VL-7B and Gemini-1.5-pro, we attribute the performance drop to increased context length and overly restrictive prompts, which limit the models' capabilities.

\subsubsection{Exploration of Intrinsic Knowledge} 
Given two challenges in VEU-Bench tasks: (1) mastering knowledge related to video editing, and (2) accurately identifying video edits, we aim to explore whether the poor performance of video language models (Vid-LLMs) is due to a lack of built-in understanding of editing concepts. We design 37 questions covering elements within these dimensions to assess the model's comprehension of editing concepts. The generated answers were evaluated by 10 volunteers, who scored each response from 1 (poor) to 3 (good). As shown in Table~\ref{tab:concept_experiment}, both the best- and worst-performing models demonstrated accurate descriptions of editing concepts. The results suggest that the poor performance of Vid-LLMs may be due to weak alignment between the intrinsic knowledge of the language model and the visual perception component. More details can be found in the Appendix.

\subsubsection{Scoring System Evaluation} We evaluate the alignment between our proposed scoring system and human preferences to assess the impact of pattern matching regularization. Ten human volunteers rated 50 answer-response pairs sampled from reasoning and judging tasks. As shown in Figure~\ref{fig:score alignment}, fewer data points concentrated below the diagonal and a higher Spearman coefficient indicate that incorporating pattern matching reduces the bias of the LLM evaluator toward scoring higher based solely on visual factual content. This results in scores that align better with human evaluations, demonstrating the robustness of our proposed evaluation system.

\section{Conclusion}
In this work, we introduce VEU-Bench, a comprehensive benchmark for evaluating Vid-LLM in the Video Editing Understanding task. Our benchmark analysis reveals significant limitations in the ability of current SOTA Vid-LLMs to comprehend and reason about video editing components. To address this, we present Oscars, a fine-tuned Vid-LLM that demonstrates substantial performance gains on VEU-Bench, exceeding leading open-sourced Vid-LLM by 28.3\% and achieving comparable performance with commercial models like GPT-4o and Gemini-pro. Furthermore, Oscars exhibits improved performance on general video understanding benchmarks, which underscores the value of VEU data to serve as the data source for enhancing the abstract reasoning ability of Vid-LLMs. VEU-Bench and Oscars provide valuable resources for advancing research in video editing understanding and enhancing the reasoning capabilities of Vid-LLMs.

{
    \small
    \bibliographystyle{ieeenat_fullname}
    \bibliography{main}
}


\end{document}